\documentclass{article}

\usepackage{microtype}
\usepackage{graphicx}
\usepackage{subfigure}
\usepackage{booktabs} 

\usepackage{hyperref}
\usepackage[hypcap=false]{caption}

\usepackage[accepted]{icml2024}

\usepackage{amsmath}
\usepackage{amssymb}
\usepackage{mathtools}
\usepackage{amsthm}

\usepackage[capitalize,noabbrev]{cleveref}

\theoremstyle{plain}

\theoremstyle{definition}

\theoremstyle{remark}

\DeclareMathOperator{\prob}{P}

\def\given{\;\middle\vert\;}

\usepackage[textsize=tiny]{todonotes}

\icmltitlerunning{Hierarchical Universal Value Function Approximators}

\begin{document}

\twocolumn[
\icmltitle{Hierarchical Universal Value Function Approximators}



\icmlsetsymbol{equal}{*}

\begin{icmlauthorlist}
\icmlauthor{Rushiv Arora}{sch,comp}
\end{icmlauthorlist}

\icmlaffiliation{sch}{University of Massachusetts, Manning College of Information and Computer Science, Amherst MA, USA}
\icmlaffiliation{comp}{Dell AI Research Office, Hopkinton MA, USA.}

\icmlcorrespondingauthor{Rushiv Arora}{rrarora@umass.edu}

\icmlkeywords{Reinforcement Learning, Universal Value Function Approximators, Hierarchical Reinforcement Learning}

\vskip 0.3in
]



\printAffiliationsAndNotice{}  

\begin{abstract}
There have been key advancements to building universal approximators for multi-goal collections of reinforcement learning value functions---key elements in estimating long-term returns of states in a parameterized manner. We extend this to hierarchical reinforcement learning, using the options framework, by introducing hierarchical universal value function approximators (H-UVFAs). This allows us to leverage the added benefits of scaling, planning, and generalization expected in temporal abstraction settings. We develop supervised and reinforcement learning methods for learning embeddings of the states, goals, options, and actions in the two hierarchical value functions: $Q(s, g, o; \theta)$ and $Q(s, g, o, a; \theta)$. Finally we demonstrate generalization of the HUVFAs and show they outperform corresponding UVFAs.

\end{abstract}

\section{Introduction}
\thispagestyle{empty}

Value functions ($V(s)$ and $Q(s, a)$) are key to reinforcement learning algorithms \cite{barto98rlbook}. They represent how important it is for the agent to be in a certain state and take certain actions in a given state. Using an optimal value function for a particular goal allows us to re-create the optimal policy the agent must follow to maximize the discounted return for that goal. Hierarchical reinforcement learning extends this to temporal abstraction \cite{sutton99smdp, precup00options} by using a hierarchy of two or more levels of value functions to create a collection of useful skills or behaviors \cite{dasilva2012skills} that solve independent tasks or subgoals. A meta-value function selects the appropriate skill in the given state while a lower-level value function drives the policy that executes the actions the agent takes when exploiting the chosen skill. This is beneficial for scaling and planning.

Universal value functions \cite{schaul15uvfa} ($V(s, g)$ and $Q(s, a, g)$) and general value functions \cite{sutton11gvf} ($V_g(s)$ and $Q_g(s, a)$) are key when extending this to multi-task domains. They incorporate information of the goal into the value function itself, making more expressive agents that can act differently in the same state as appropriate for the current goal. A collection of universal and general value functions over a diverse set of goals is key in building a single unified expression for generating policies and building agents with multiple capabilities (multi-task capabilities).

Complex environments require parameterized representations of the value functions ($V(s; \theta)$ or $Q(s, a; \theta)$), often with large neural networks. These parameterized representations exploit the structure of the state space to learn a mapping to the value functions and generalize behavior to unseen states. Universal value function approximators \cite{schaul15uvfa} demonstrated that one can learn parameterized representations of \textit{universal} value functions $V(s, g; \theta)$ that exploit the underlying structure of the goal and state space to generalize to unseen goals and states. They exploit the universal function approximation capabilities of neural networks to learn a single value function that represents states and goals.

We extend UVFAs to hierarchical reinforcement learning, or temporal abstraction, using the options framework. Since there are two or more levels of value functions in a hierarchy, we must learn representations (both parameterized and non-parameterized) of all the levels of the hierarchy to result in a final state of value functions equal in number to the hierarchy levels. The challenge lies in extending the concept of UVFAs from two or three dimensions (for the state and action value functions respectively) to \emph{n}-dimensions. Our method is novel because it shows that there is underlying structure in the states, goals, \emph{options and actions} that results in a universal representation of the hierarchy. This universal representation results in universal temporally abstract behaviors or skills that has \emph{zero-shot} generalization to unseen goals. We show that this representation outperforms corresponding UVFAs.


\section{Background}

A Markov Decision Process (MDP) $(\mathcal{S}, \mathcal{A}, \mathcal{T}, \mathcal{R}, \gamma)$ consists of a set of states $\mathcal{S}$, a set of actions $\mathcal{A}$, a transition function $\mathcal{T}(s, a, s') \coloneq \prob(\mathcal{S}_{t+1}=s' | \mathcal{S}_t=s, \mathcal{A}_t=a)$, a reward function $\mathcal{R} : \mathcal{S} \times \mathcal{A} \to \mathbb{R}$ and a reward \textit{discount factor} $\gamma \in [0, 1)$. Since we consider a multi-task setting, each goal has its own reward function $\mathcal{R}_g$ and discount factors $\gamma_g$. A Markovian \textit{policy} $\pi: \mathcal{S} \times \mathcal{A} \to [0,1] \coloneq \prob (\mathcal{A}_t=a | \mathcal{S}_t=s)$ is a probability distribution of all the actions conditioned over the states. The agent's goal is to construct a policy that maximises the state value function $V_\pi(s)$ and action-value function $Q_\pi(s, a)$ defined as

$$V_\pi(s) = \mathbb{E}_{\pi}\left[ \sum_{t=0}^\infty \gamma^{t} \mathcal{R}_{t+1} \given s_0=s \right]$$

$$Q_\pi(s, a) = \mathbb{E}_{\pi}\left[ \sum_{t=0}^\infty \gamma^{t} \mathcal{R}_{t+1} \given s_0=s, a_0 = a \right]$$

Solving for the optimal action-value function allows one to deduce a greedy optimal policy. For discrete MDPs there is at least one optimal policy that is greedy with respect to its action-value function.

\textbf{Universal Value Functions} \cite{schaul15uvfa} extend value functions to include the goals and produce the universal state value function $V_\pi(s, g)$ and universal action value function $Q_\pi (s, a, g)$. These universal value functions are defined in the context of multi-task learning over multiple goals $\mathcal{G}$ and are constructed using multiple individual value functions $V_{g,\pi}(s)$ and $Q_{g,\pi}(s, a)$ defined for each goal $g \in \mathcal{G}$. Universal value functions can be learned using function approximators to become universal value function approximators (UVFAs). Given UVFAs that are trained on a diverse set of goals and parameterized with sufficient capacity, we learn an underlying structure of the states and goals that can be used to construct value functions that extend to unseen states or goals.

\textbf{The Options Framework} \cite{sutton99smdp, precup00options} formalizes the concept of temporal abstractions in reinforcement learning, or hierarchical reinforcement learning, by using semi-MDPs. A Markovian option $o \in \Omega$ is a tuple $(\mathcal{I}_{o}, \pi_{o}, \beta_{o})$ where $\mathcal{I}_{o} \in \mathcal{S}$ is its initiation set, $\pi_{o}$ is its intra-option policy, and $\beta_{o}: \mathcal{S} \to [0, 1]$ is its termination function. We assume that $\forall s \in \mathcal{S}$, $\forall o \in \Omega$: $s \in \mathcal{I}_{o}$--all options are available everywhere. An option $o$ is picked from the initiation set $\mathcal{I}_o$ using the policy-over-options $\pi_\Omega(s, o)$ and follows its intra-option policy $\pi_o(s, a)$ until it terminates by $\beta_o(s)$ which determines the probability of a given option terminating in a given state.

This is a call-and-return model where the option initiation is determined by the policy-over-options (meta-policy) $\pi_\Omega(o | s )$ that gives us the probability of selecting option $o$ in state $s$ and is driven by the option-value function $Q_\Omega(s, o): \mathcal{S} \times \Omega \to \mathbb{R}$. Once the option is selected, it performs actions using the intra-option policy $\pi_U(a | s, o)$ which is driven by its intra-option value function $Q_U(s, o, a): \mathcal{S} \times \Omega \times \mathcal{A \to \mathbb{R}}$. This continues until the option terminates, upon which the process restarts in the then-current state. The two sets of value functions are related as

\begin{align}
Q_\Omega(s, o)  =\sum_a \pi_{o, \theta}\left( a \given s \right)Q_U(s, o, a)\enspace , \label{eq:q_omega}
\end{align}

\begin{align}
Q_U(s, o, a) = r(s,a) + \gamma \sum_{s'} P \left( s' \given s, a \right) U(o, s') \enspace . \label{eq:q_u}
\end{align}

where
\begin{align*}
U(o, s') = (1 - \beta_{o, \vartheta}(s')) Q_\Omega(s', o) + \beta_{o, \vartheta}(s') V_\Omega(s')  \enspace
\end{align*}

is the value of executing $o$ upon entering a state $s'$.

\section{Hierarchical Universal Value Function Approximators}

Hierarchical universal value function approximators \textit{(H-UVFAs)} extend UVFAs to hierarchical reinforcement learning to give zero-shot generalization to unseen goals by exploiting underlying structure in the states, goals, \emph{options and actions} that results in a universal representation of the hierarchy. We aim to construct H-UVFAs over every level in the hierarchy (in our case, we use a two level hierarchy). Particularly, we aim to construct $Q_\Omega(s, o, g; \theta)$ and $Q_U(s, o, a, g; \eta)$ where the H-UVFA meta-policy is parameterised by $\theta \in \mathbb{R}^m$ and the H-UVFA intra-option policy is parameterized by $\eta \in \mathbb{R}^n$. Note that $m$ and $n$ don't necessarily have to be equal. These H-UVFAs draw upon hierarchical general value functions ($Q_\Omega(s, o, g; \theta) \approx Q_{\Omega, g}^*(s, o)$ and $Q_U(s, o, a, g; \eta) \approx Q_{U, g}^*(s, o, a)$) which, to the best of our knowledge, haven't been defined before and will be created in the process of building H-UVFAs.

\begin{figure}[ht]
\vskip 0.2in
\begin{center}
\centerline{\includegraphics[width=\columnwidth]{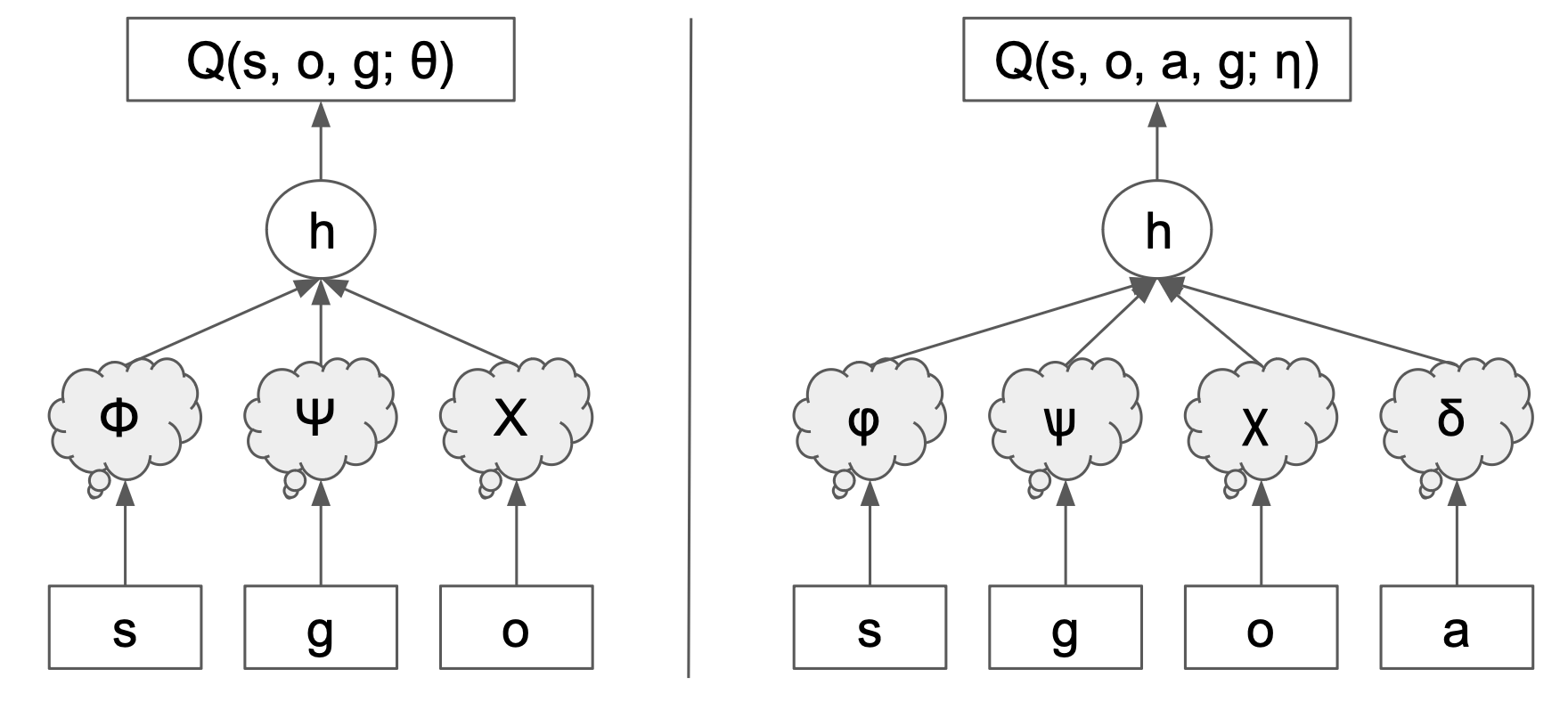}}
\caption{The \textit{multi-stream} architecture of H-UVFAs. In different temporal abstraction methods with more hierarchies, there will be more streams. This will result in high-order dimensionality reductions, which our method can accommodate as we will see in the upcoming sections. \textbf{Left:} The three-stream architecture of the policy-over-options (meta-policy) H-UVFA. The streams are: states, goals, options. \textbf{Right:} The four-stream architecture of the intra-option policy H-UVFA. The streams are: states, goals, options, actions.}
\label{fig:two-stream}
\end{center}
\vskip -0.2in
\end{figure}

Like UVFAs, we propose learning by using a \textit{multi-stream} architecture as seen in Figure \ref{fig:two-stream}. Each component of the value function gets its own stream of embeddings.
\begin{itemize}
    \item The meta-H-UVFA $Q_\Omega(s, o, g; \theta)$ can be split into three sets of embeddings: $\Phi: \mathcal{S} \to \mathbb{R}^m$, $\Psi: \mathcal{G} \to \mathbb{R}^m$ and $X: \Omega \to \mathbb{R}^m$
    \item The intra-option H-UVFA $Q_U(s, o, a, g; \eta)$ can be split into four sets of embeddings: $\phi: \mathcal{S} \to \mathbb{R}^n$, $\psi: \mathcal{G} \to \mathbb{R}^n$, $\chi: \Omega \to \mathbb{R}^n$, and $\delta: \mathcal{A} \to \mathbb{R}^n$
\end{itemize}

Notice that we use upper-case Greek letters for the higher level meta-policy and lower-case Greek letters for the lower level intra-option policy.

Building H-UVFAs is a two-step process.
\begin{itemize}
    \item First, we need to construct the hierarchical universal value functions. That is, we need to construct the $m-$ or $n-$ dimensional embeddings for the different streams. In our methods, we will construct a high-order tensor with the number of dimensions equal to the number of streams in the desired H-UVFA (different depending on the level) and decompose it into the different embeddings.
    \item Next, we need to construct the hierarchical universal function \textit{approximators}. That is, we need to train a sufficiently parameterized approximator such as a neural network (universal function approximator) on the embeddings.

    The results of the H-UVFAs are passed through $h(.)$ that operates on the embeddings and converts it into a scalar. In our method, as described ahead, this is the dot product of all the embeddings resulting from the approximators.
\end{itemize}

The first step is equivalent to a higher-order singular value decomposition (SVD) on two-dimensional matrices, except done on higher-dimension tensors. We achieve this decomposition into the $m-$ and $n-$ dimension embeddings using the PARAFAC decomposition \cite{tomasi05parafac, kolda09parafac}. Alternatively, one can use the Tucker decomposition \cite{tucker66decomposition}. If we assume that the given H-UVFA has $N$ streams, then we know that the higher-order tensor will have $N > 2$ dimensions and the result will be $N$ sets of $m-$ or $n-$ dimensional embeddings

The second step is a simple function approximation task. We can use any function approximator and training method to approximate the resulting collection of embeddings. For $N$ streams, we will use $N$ separate approximators. That is, we use one network each for each stream---as an example, the meta-policy H-UVFA has three approximators and the intra-option policy H-UVFA has four approximators.

Notice that the meta-policy H-UVFA $Q_\Omega(s, o, g; \theta)$ is equivalent to the UVFA for single-policy reinforcement learning $Q(s, a, g; \theta)$ except we use options instead of actions. However, the experiments in \cite{schaul15uvfa} are based on a two stream architecture and use SVD or OptSpace \cite{keshavan09optspace} while our experiments describe higher-order stream architectures and use PARAFAC for decomposition.

\section{Supervised Learning of H-UVFAs}

\subsection{Method}

The supervised learning approach to building H-UVFAs is best applicable when it is possible to \textit{iterate} over the states, goals, options, and actions to construct the matrix that we can decompose into embeddings.

The two-step multi-stream process described in the previous section is achieved using the following steps to learn the parameters $\theta$ and $\eta$:
\begin{itemize}
    \item Iterate over all the states, goals, actions, and options and organize their $Q$-values into two multi-dimensional tensors. The first tensor represents $Q_\Omega(s, o, g)$ and is a three dimensional tensor with one dimension each for the states, goals, and options. The second tensor is the four dimensional $Q_U(s, o, a, g)$ with one dimension each for the states, goals, options, and actions. The tensors are of dimensions $|\mathcal{S}| \times |\Omega| \times |\mathcal{G}|$ and $|\mathcal{S}| \times |\Omega| \times |\mathcal{A}| \times |\mathcal{G}|$ respectively.
    \item Decompose the two multi-dimensional tensors into a set of embedding \textit{vectors} corresponding to each dimension using either PARAFAC \cite{kolda09parafac} or Tucker decompositions \cite{tucker66decomposition}. As discussed before, the embedding \textit{vectors} correspond as $Q_\Omega: (s, g, o) \to  (\hat{\Phi},  \hat{\Psi}, \hat{X})$ and $Q_U: (s, g, o, a) \to (\hat{\phi}, \hat{\psi}, \hat{\chi}, \hat{\delta})$.
    \item Train a function approximator to approximate the embeddings from the embedding vectors to learn the parameters $\theta = \{\Phi, \Psi, X \}$ and $\eta = \{\phi, \psi, \chi, \delta \}$. This now results in $Q_\Omega(s, o, g; \theta)$ and $Q_U(s, o, a, g; \eta)$.
\end{itemize}

This method minimizes the mean-squared-errors (MSE) $\mbox{\textit{MSE}}_{Q_\Omega} = [Q_{\Omega, g}(s, o) - Q_\Omega(s, o, g; \theta) ]^2$ and $\mbox{\textit{MSE}}_{Q_U} = [Q_{U, g}(s, o, a) - Q_U(s, o, a, g; \eta) ]^2$

The first and second step result in the creation of H-\textit{UVAs} $Q_\Omega(s, o, g)$ and $Q_U(s, o, a, g)$. These unparameterized tensors and their embedding vectors are useful when the agent's policy needs to be inferred for goals that have been observed when collecting the data. The function approximators that learn embeddings from the embedding vectors construct the H-\textit{UVFAs} from the H-\textit{UVFs}---the idea being that similar state, goal, option, and action pairs will have similar values to the corresponding pairs in the observed data. We will see in the experiments that the factorization into the respective embeddings will exploit the underlying structure, thereby resulting in generalization.

\subsection{Supervised Learning Experiments}

\textbf{Setting:} We construct H-UVFAs in the four rooms domain \cite{sutton99smdp} with the goals spread out over three rooms--top left, top right, and bottom-left--with the fourth room being used to test generalization of the H-UVFAs. We allow the agent to be randomly initialized in any of the four rooms at the start of each episode in order to ensure that it can gain a representation of all rooms. The agent receives a reward of 1 for reaching the goal and a reward of 0 otherwise. All episodes terminate after 1000 timesteps. More details for the experiment can be found in Appendix \ref{experiment_details_appendix}

\begin{figure}[!ht]
\vskip 0.2in
\begin{center}
\centerline{\includegraphics[width=\columnwidth]{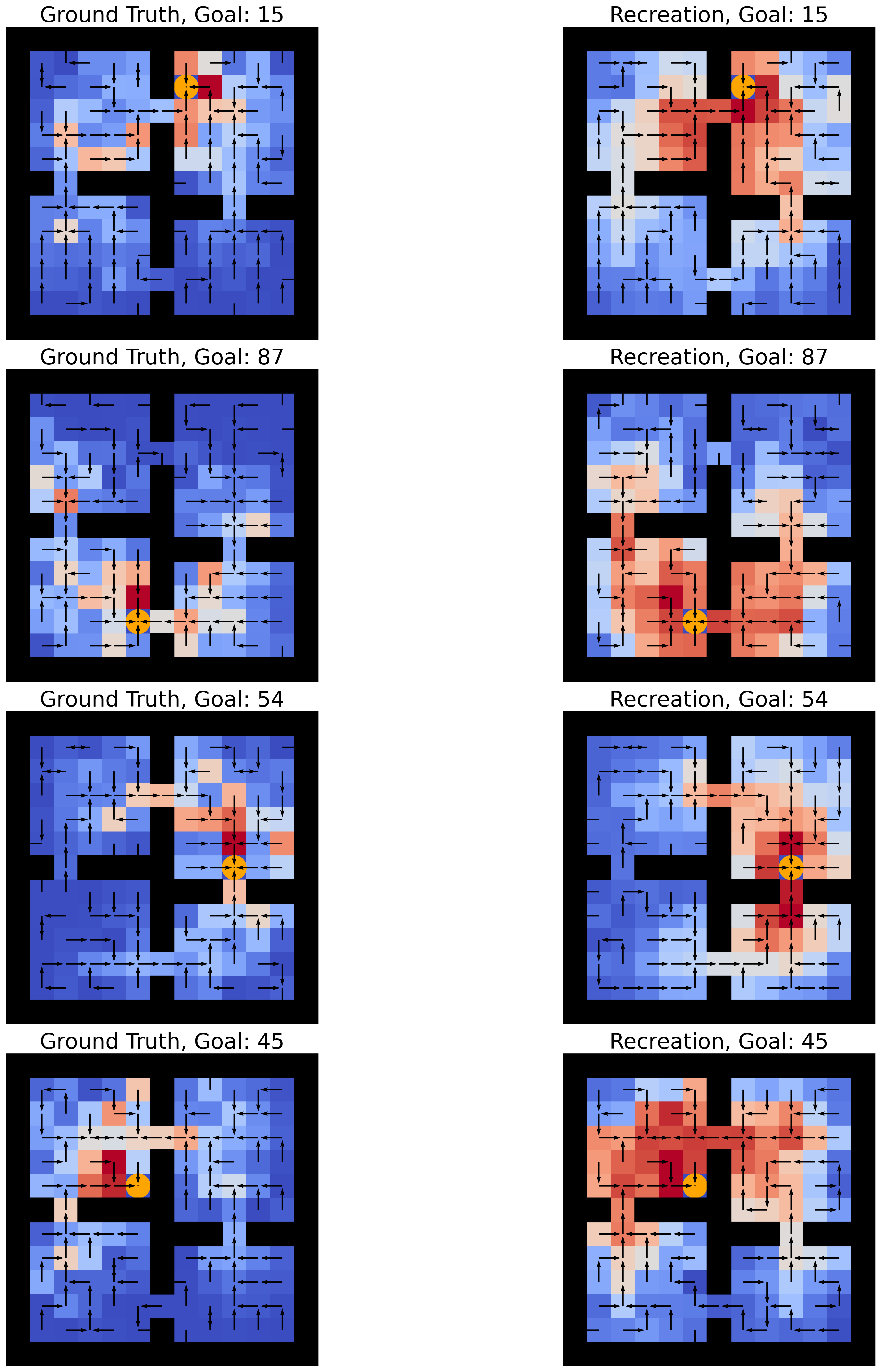}}
\caption{A comparison of the values and actions of the ground truth and H-UVFAs. Red represents higher values for each state, blue represents lower values of being in a state, and the orange circle represents the goal state. The arrows represent the greedy action in each state obtained from picking a greedy option. It is interesting to note that using neural networks/function approximators (H-UVFAs) allows for gradual and smooth changes in the states' values as compared to the more drastic ones seen in the discrete and tabular ground truth. This better represents policies and their values and allows for better agent behaviors in long-trajectories where the agent is far from the goal.}
\label{fig:action-recreation}
\end{center}
\vskip -0.2in
\end{figure}

\textbf{Results:} We construct H-UVFAs using a combination of 25 goals. Figure \ref{fig:action-recreation} plots the values and greedy actions of the H-UVFAs on a subset of the goals used during training along with the ground truth values corresponding to the option-value and intra-option value functions. In each state, the greedy option is picked and the corresponding value and greedy action are plotted. We find that an embedding rank of 50 is sufficient to re-create the ground truth matrix accurately. We plot the entire decomposed matrices and the corresponding re-created H-UVFAs for $Q_\Omega$ and $Q_U$ in Appendix \ref{matrix_recreations}.

Interestingly, we see that the re-creation of the value function with a neural network allows for more continuous changes in the values---as we move further away from the goal the value of the state decreases gradually as compared to the more sudden changes in value functions seen in the discrete and tabular ground truth. This behavior better represents a value function and policy and allows for better behavior in longer trajectories where more optimal actions are picked further away from the goal as seen in Figure \ref{fig:action-recreation}.

\begin{figure}[!ht]
\vskip 0.2in
\begin{center}
\centerline{\includegraphics[width=\columnwidth]{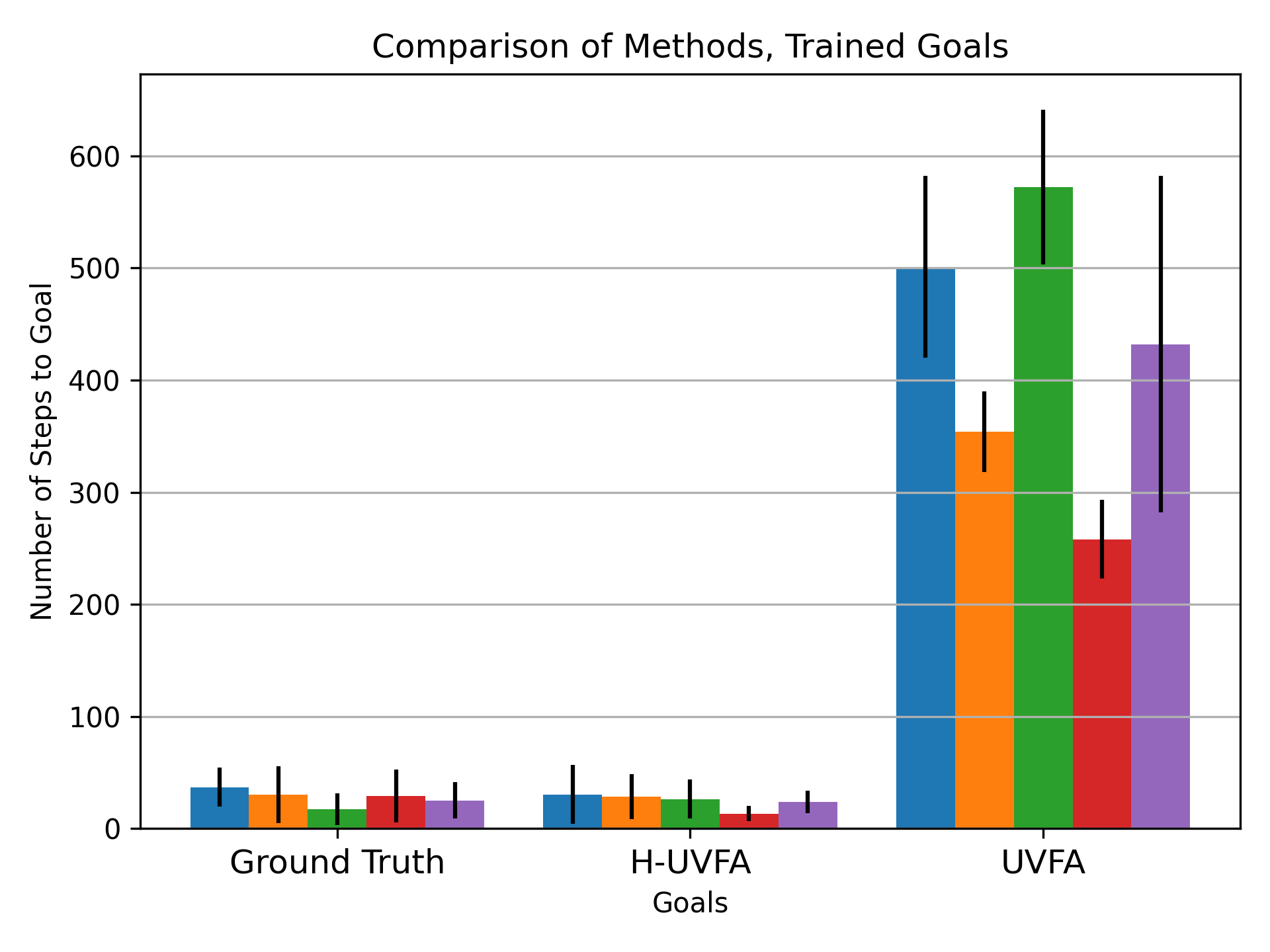}}
\caption{Comparison of the ground truth, H-UVFAs, and UVFAs on goals seen during training as measured in averages and standard deviations over 10 episodes for 5 goals. H-UVFAs and ground truth are comparable in performance while UVFAs have poor performance and higher variance. The poor performance of UVFAs as compared to H-UVFAs in hierarchical settings is is due to the loss of information when a large amount of information is compressed in a compact representation, as discussed in Section \ref{uvfa-comparison}.}
\label{fig:ground-truth-comparison-supervised-learning}
\end{center}
\vskip -0.2in
\end{figure}

Figure \ref{fig:ground-truth-comparison-supervised-learning} plots the comparison of the ground truth, H-UVFAs, and UVFAs on 5 trained goals as represented by the average steps to goal and its standard deviation measured over 10 episodes for each goal. H-UVFAs and the ground truth H-UVF matrix have comparable performance. It is possible to construct UVFAs which provide better performance than an agent acting randomly; however, we notice significant decrease in performance and higher variance for the base method. This is due to the fact that, in hierarchical settings, the lower-level UVFAs pack significant behavior in a compact representation that doesn't generalize over different skills. This is further discussed in Section \ref{uvfa-comparison}.

\begin{figure}[ht]
\vskip 0.2in
\begin{center}
\centerline{\includegraphics[width=\columnwidth]{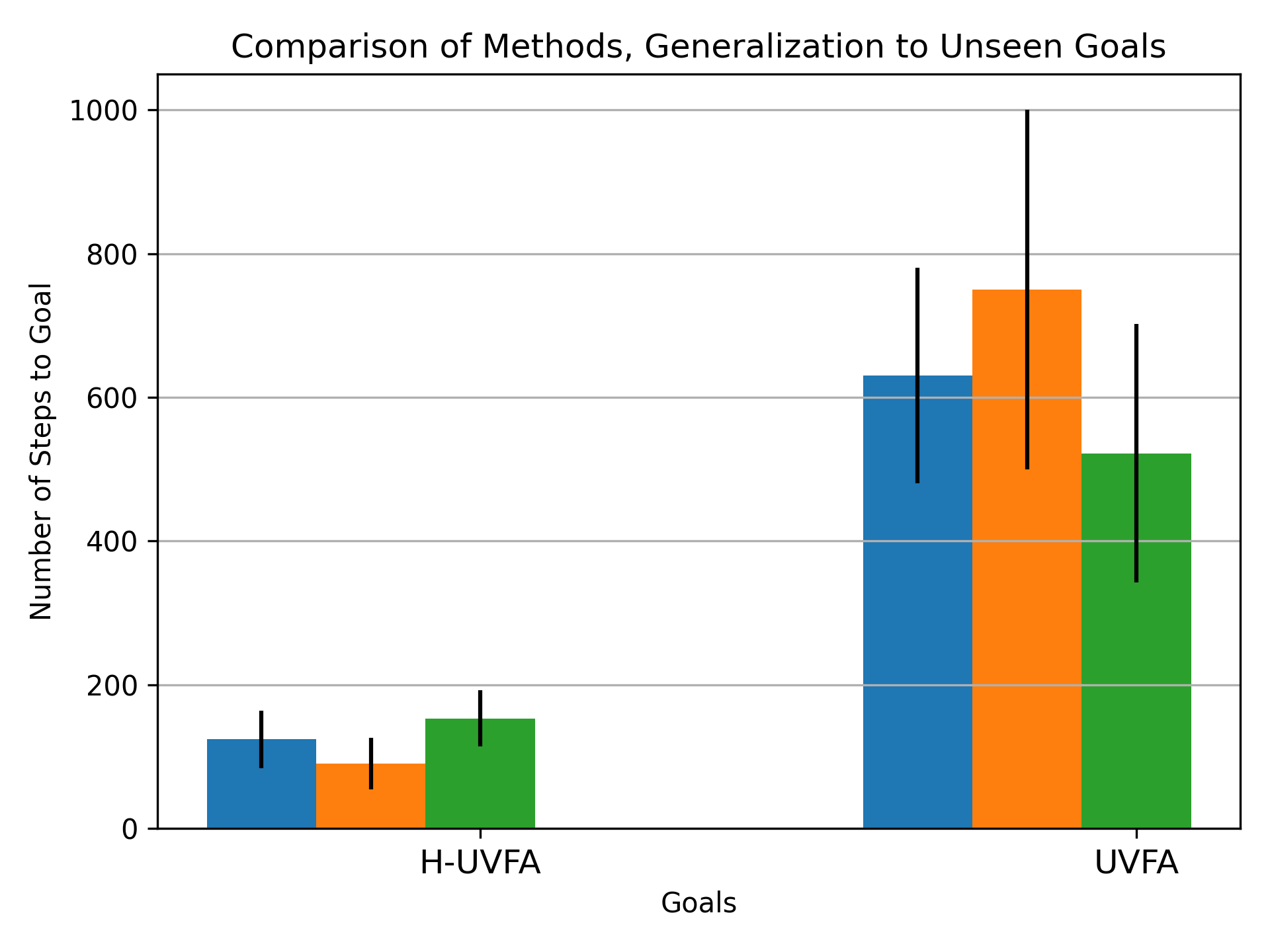}}
\caption{Comparison of H-UVFAs and the UVFAs baseline in generalization to unseen goals as measured in average steps to goal and standard deviation over 10 episodes each for 3 goals. H-UVFAs generalize well in hierarchical settings to unseen goals. The variance for both methods increases as compared to that on trained goals. }
\label{fig:comparison-supervised-learning-generalization}
\end{center}
\vskip -0.2in
\end{figure}

\begin{figure}[ht]
\vskip 0.2in
\begin{center}
\centerline{\includegraphics[width=\columnwidth]{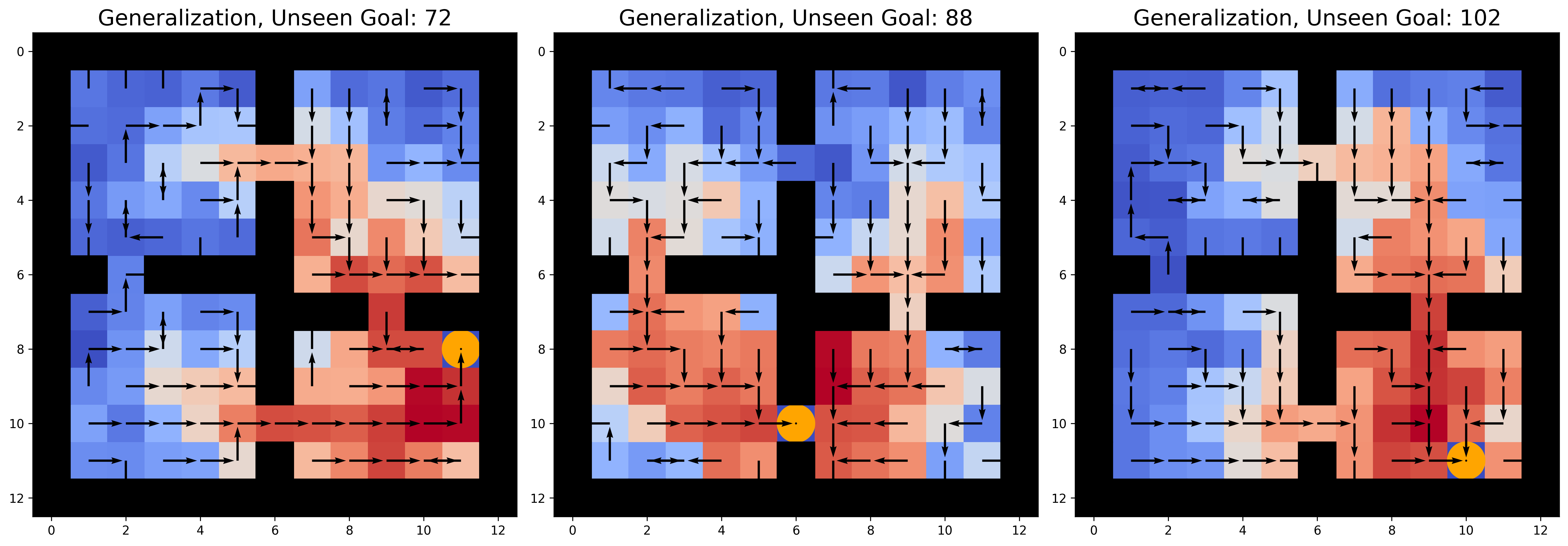}}
\caption{Values and policies for unseen goals in the fourth room. Red indicates higher values while the arrows indicate the greedy action for the greedy policy. The near-optimal value functions and policies indicate that H-UVFAs can extrapolate to create optimal hierarchical behaviors.}
\label{fig:action-recreation-generalization}
\end{center}
\vskip -0.2in
\end{figure}

\textbf{Generalization:} We measure generalization of the H-UVFAs to unseen rooms by constructing similar plots, as seen in Figures \ref{fig:comparison-supervised-learning-generalization} and \ref{fig:action-recreation-generalization}, for goals in the fourth room. We note a similar continuous value function and optimal actions for unseen goals. We also note that when measured over multiple episodes and unseen goals H-UVFAs still perform well and reach the goal in an optimal number of steps. This generalization to unseen goals reveals that the H-UVFAs are able to exploit the underlying structure in the states, goals, options, and actions to generate embeddings for each that can be combined to produce value functions and policies that extrapolate. This zero-shot generalization and better performance than the vanilla UVFAs baseline indicates that universal and general value functions can be created in hierarchical settings.

\section{Reinforcement Learning of H-UVFAs}

We now present results in the more practical setting of reinforcement learning. This method is applicable to continuous domains where it is not possible to access the value functions for every possible combination of states, goals, options, and actions. We find that it is possible to construct H-UVFAs simply using the observed values of the value functions seen during episodes. We will continue to build on the multi-stream architecture we used in supervised learning settings. Like most hierarchical reinforcement learning approaches, we assume a finite set of options.

\subsection{Method}

We build upon the method presented by \citet{schaul15uvfa} to construct two methods of developing H-UVFAs: using a Horde \cite{sutton11gvf} of general value functions, and directly using bootstrapping.

The first method using a Horde is depicted in Algorithm \ref{alg:huvfa-horde}. We learn multiple policies on a diverse set of goals---a Horde---and construct hierarchical general value functions $Q_{\Omega, g}$ and $Q_{U, g}$. Using these hierarchical general value functions, we build two data matrices that we factorize into the respective state, goal, option, and action embeddings that the H-UVFAs approximate. To the best of our knowledge, no other work introduces hierarchical general value functions (H-GVFs) that we use in constructing our H-UVFAs.

Algorithm \ref{alg:huvfa-horde} constructs the H-GVFs and H-UVFAs using a series of sample episodes obtained from the Horde. The transitions in all the sample episodes can be used to construct the data matrices $M_{\Omega; t, g}$ and $N_{U; t, g}$, for the option-value and intra-option value functions respectively. Unlike UVFAs where each row represents time-index of one transition in the history and each column corresponds to one goal, the data matrices in this process can have more than one dimension:
\begin{itemize}
    \item For the option-value function $Q_{\Omega}$ we can expect a three dimensional matrix where the first dimension corresponds to the time, the second to the number of goals, and the third to the options. We can achieve this since we assume that the number of options is finite. 
    
    In more likely cases it is also possible to do something similar to UVFAs and simply use the value of the observed option instead of all possible options, in which case we would obtain a two-dimensional tensor with the factorized streams being over $\Phi(s, o)$ and $\Psi(g)$ instead of $\Psi(s)$, $\Psi(g)$ and $X(o)$
    \item For the intra-option value function $Q_U$ we can expect three or four dimensions. The first dimension corresponds to the time-index of one transition in the history, the second dimension corresponds to each goal in the Horde. In the first variant the third dimension corresponds to the number of options, with the value for the observed action being stored for all options---a case more likely with infinite actions. In the second variant with a finite number of actions, the third dimension can correspond to the options and the fourth to the actions.
\end{itemize}

\textit{Since we would like to construct H-UVFAs using only our transition history and no additional data or assumptions}, we conduct experiments using a two-dimensional data matrix for $Q_\Omega$ and a three-dimensional data matrix for $Q_U$.

The challenge in this method arises from the likely scenario of not observing enough transitions that cover the entire state, goal, option, and action space sufficiently. We find that with enough transitions and a diverse set of target in the Horde, the method is sufficient to learn H-UVFAs. Similar to the results for the supervised learning experiments, these H-UVFAs generalize to unseen goals and outperform UVFAs on both the trained and unseen goals.

\begin{algorithm}[ht]
   \caption{H-UVFAs from Horde}
   \label{alg:huvfa-horde}
\begin{algorithmic}
   \STATE {\bfseries Input:} rank $m$, rank $n$, training goals $\mathcal{G}$, budgets $b_1$, $b_2$, $b_3$
    \STATE Initialize transition history $\mathcal{H}$
    \FOR{$t=1$ {\bfseries to} $b_1$}
    \STATE $\mathcal{H} \leftarrow \mathcal{H} \cup (s_t, o_t, a_t, s_{t+1})$
    \ENDFOR
    \FOR{$i=1$ {\bfseries to} $b_2$}
    \STATE Pick a random transition in $\mathcal{H}$
    \STATE Pick a random goal (or iterate over goals)
    \STATE Update $Q_{\Omega, g}$
    \STATE Update $Q_{U, g}$
    \ENDFOR
    \STATE Initialize data matrices $M_{\Omega; t, g}$ and $N_{U; t, g}$
    \FOR{$(s_t, o_t, a_t, s_{t+1})$ {in} $\mathcal{H}$}
    \FOR{$g \in \mathcal{G}$}
    \STATE $M_{\Omega; t, g} \leftarrow Q_{\Omega, g}(s_t, o_t)$
    \STATE $N_{U; t, g} \leftarrow Q_{U, g}(s_t, o_t, a_t)$
    \ENDFOR
    \ENDFOR
    \STATE Decompose into embeddings $Q_\Omega(s, o, g) = \Phi.\Psi.X$ and $Q_U(s, o, a, g) = \phi.\psi.\chi.\delta$
    \STATE Initialize three function approximators (neural networks in our case) for $Q_\Omega(s, o, g; \theta)$ and four function approximators for $Q_U(s, o, a, g; \eta)$
    \FOR{$i = 1$ {\bfseries to} $b_3$}
    \STATE Learn parameters $\theta$ and $\eta$ by:
    \STATE Updating $\hat{\Phi}(s_t)$ to $\Phi$, $\hat{\Psi}(g_t)$ to $\Psi$ and $\hat{X}(o_t)$ to $X$
    \STATE Updating $\hat{\phi}(s_t)$ to $\phi$, $\hat{\psi}(g_t)$ to $\psi$, $\hat{\chi}(o_t)$ to $\chi$ and $\hat{\delta}(a_t)$ to $\delta$
    \ENDFOR
    \STATE \textbf{return} $Q_\Omega(s, o, g;\theta) \coloneq h(\Phi(s), \Psi(g), X(o))$ and $Q_U(s, o, a, g; \eta) \coloneq h(\phi(s), \psi(g), \chi(o), \delta(a))$
\end{algorithmic}
\end{algorithm}

\subsection{Reinforcement Learning Experiments}

\subsubsection{Horde Generalization}

The first set of reinforcement learning experiments construct H-UVFAs using a target of Hordes as described in Algorithm \ref{alg:huvfa-horde}. We use a target of 15 diverse Hordes, consisting of 15 goals spread out over three of the four rooms. The transition history is collected using two complete episodes from each Horde and the data matrix is populated using only the transition history as described in the previous section. Like the supervised learning experiments, we use a rank of 50 for factorizing the embeddings both the value functions.

\begin{figure}[h]
\vskip 0.2in
\begin{center}
\centerline{\includegraphics[width=\columnwidth]{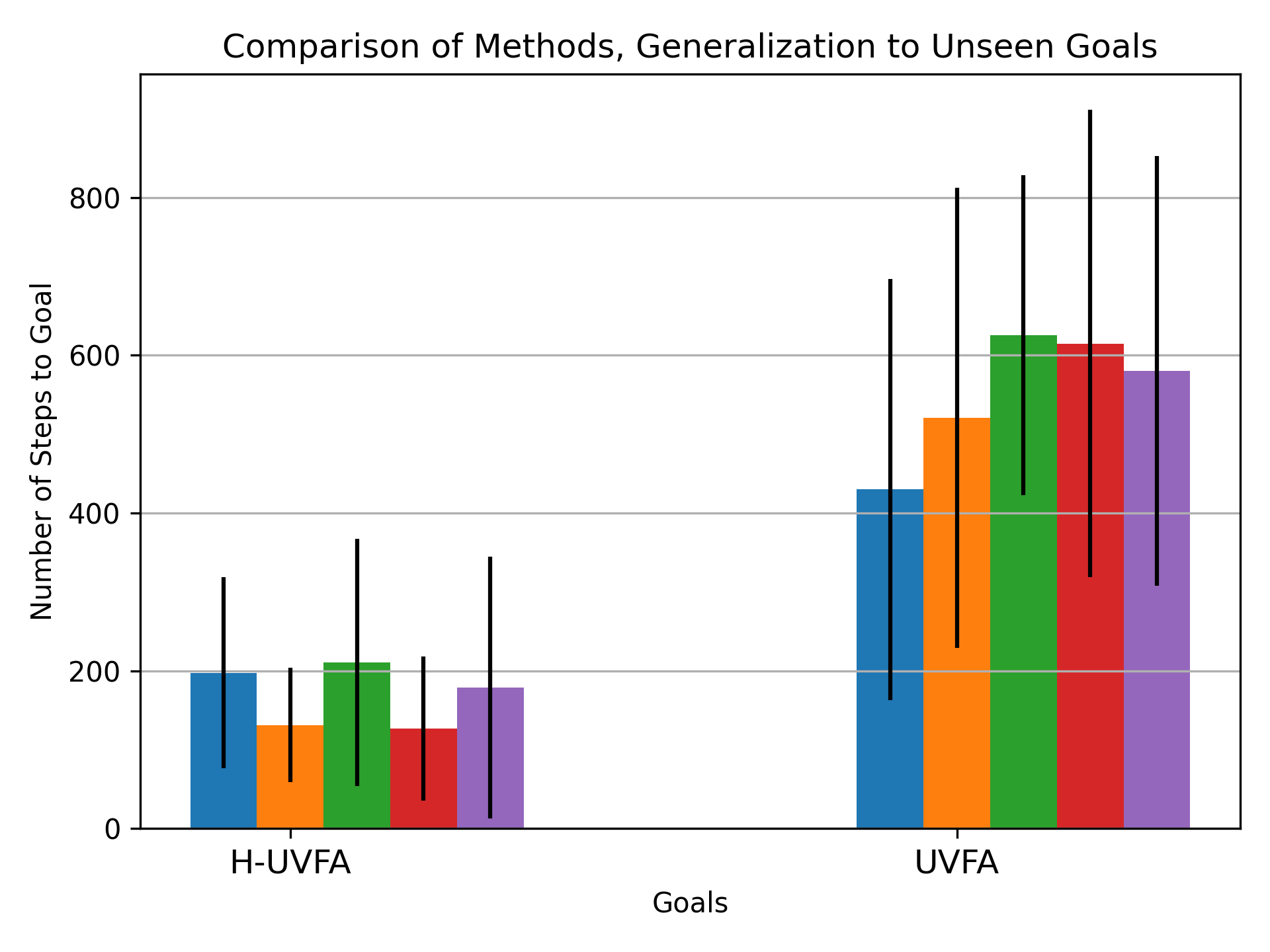}}
\caption{Comparison of H-UVFAs and UVFAs to unseen goals when trained with reinforcement learning from a Horde of targets. H-UVFAs still outperform UVFAs. This method has higher variance.}
\label{fig:comparison-reinforcement-learning-generalization}
\end{center}
\vskip -0.2in
\end{figure}

We test extrapolation of the learned H-UVFAs to goals in the fourth room and plot the results in Figure \ref{fig:comparison-reinforcement-learning-generalization}. The agent guided by H-UVFAs is able to perform well and generalize to unseen goals; however, this is not the case when using UVFAs. This comparison in performance is further discussed in Section \ref{uvfa-comparison}. It is possible, in both the supervised and reinforcement learning experiments, to achieve similar performance with a lower rank. 

We find that the reinforcement learning approach has slightly higher variance, as expected due to the stochastic conditions of data collection.

\textbf{Atari Experiments: } Like \citet{schaul15uvfa}, we scale our method for learning H-UVFAs to the Arcade Learning Environment \cite{bellemare13arcade} with the Ms. Pacman domain. The training goal set for the agents in the Horde consists of 30 pellets selected randomly throughout the space; the remaining 120 pellets are test goals used to measure generalization of the H-UVFAs. Each pellet is the only visible goal to the agent with all the other pellets being masked. The reward function is modified to provide a positive reward only upon reaching the correct pellet, with no reward being provided for any other pellet. Further experiment details can be found in Appendix \ref{experiment_details_appendix}.

Figure \ref{fig:atari-results} plots the distribution of the training and test pellets as well as the ground truth and H-UVFA retrieved value functions for three test goals. The H-UVFAs learned using the smaller Horde of agents can accurately approximate the value-functions of the unseen pellets. The policy learned from following the H-UVFAs is comparable to the ground truth policy due to the accuracy in the value function re-creation. H-UVFAs are able to exploit the underlying structure in the states, goals, options, and actions in higher dimensional spaces.

\begin{figure}[h]
\vskip 0.2in
\begin{center}
\centerline{\includegraphics[width=0.75\columnwidth]{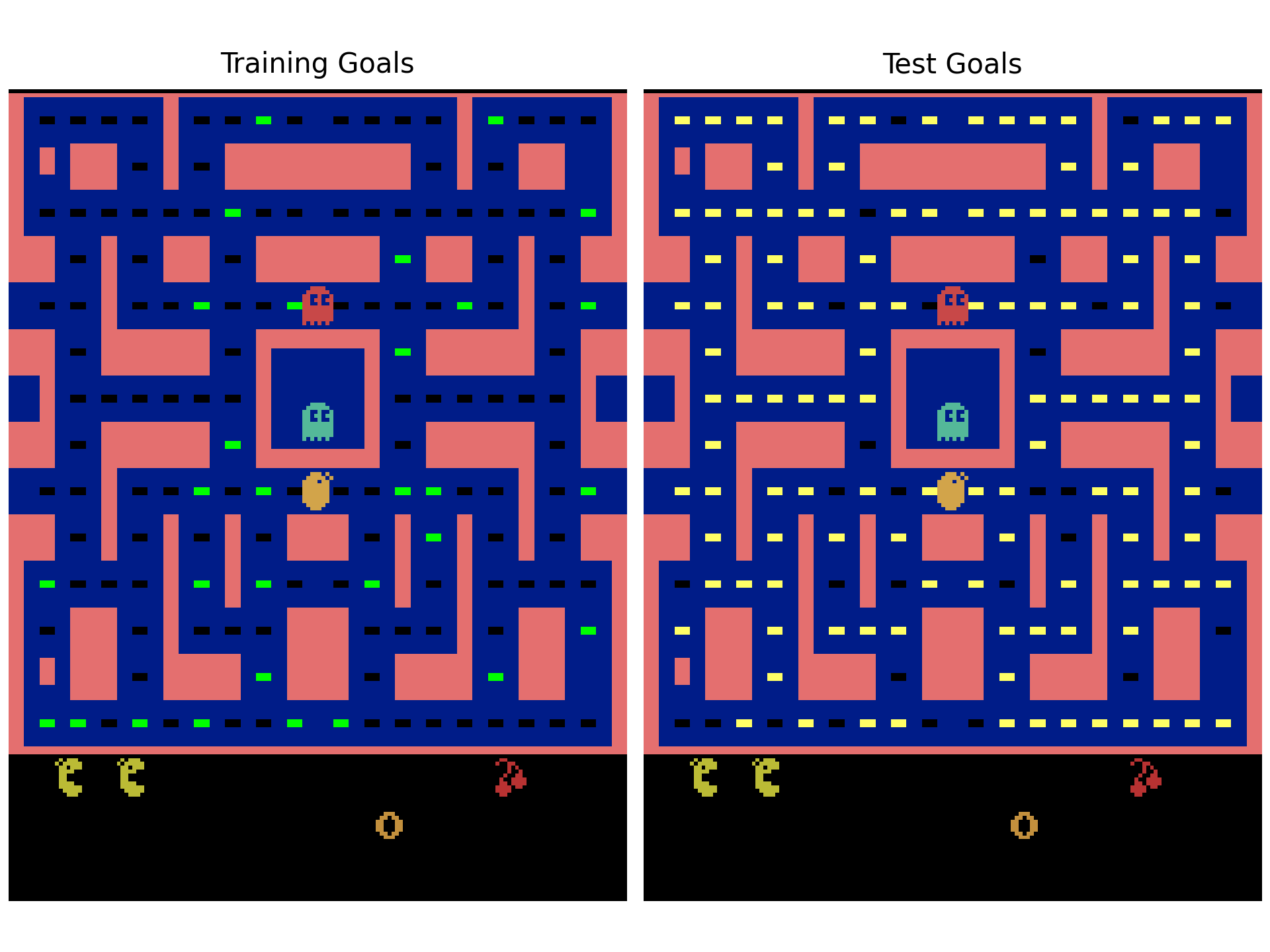}}
\centerline{\includegraphics[width=\columnwidth]{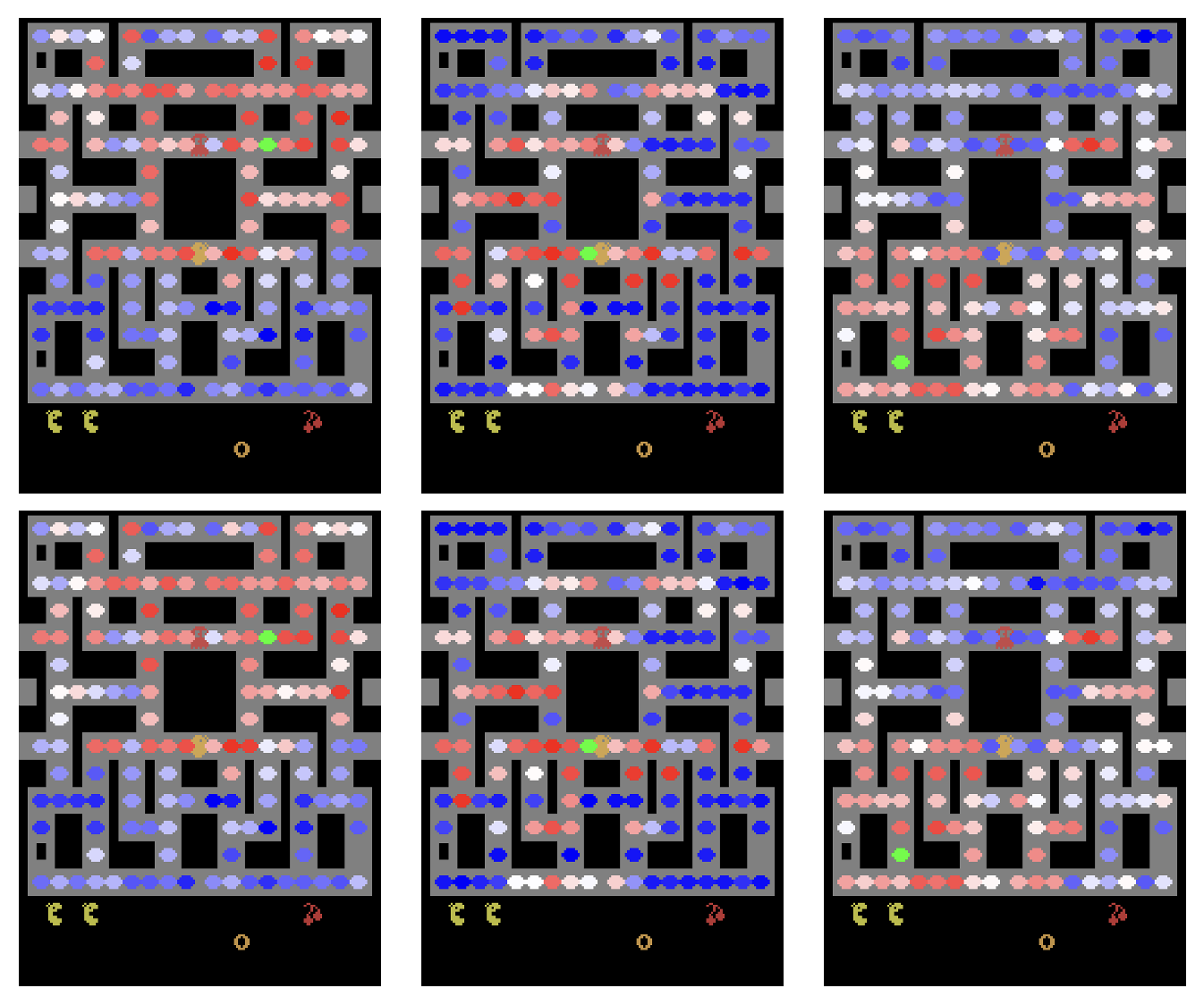}}
\caption{H-UVFA results on higher dimensional Ms. Pacman Domain. The top row plots the training and test pellet distribution in green and yellow respectively. The center and bottom rows depict the ground truth and H-UVFA re-created value functions respectively on test goals. Higher values are indicated in red, lower-values are indicated in blue, and the test goal is indicated in green. H-UVFAs can exploit the underlying structure in states, goals, options, and actions in higher dimensions as seen by their generalization when re-creating value functions for unseen goals.}
\label{fig:atari-results}
\end{center}
\vskip -0.2in
\end{figure}

\subsubsection{Learning with Bootstrapping}

Similar to UVFAs, it is possible to learn H-UVFAs directly using bootstrapping following value function updates:

\begin{multline*}
Q_\Omega(s_t, o_t, g) \leftarrow \alpha \Bigl[ r_t + \gamma \Bigl( (1 - \beta_{o_t}(s_{t+1}))Q_\Omega(s_{t+1}, o_t, g) \\ + \beta_{o_t}(s_{t+1}) \max\limits_{o'} Q_\Omega(s_{t+1}, o', g) \Bigr) \Bigr] + (1 - \alpha)Q_\Omega(s_t, o_t, g)
\end{multline*}

and 

\begin{multline*}
    Q_U(s_t, o_t, a_t, g) \leftarrow \alpha \Bigl[ r_t + \\ \gamma \Bigl( (1 - \beta_{o_t}(s_{t+1})) Q_U(s_{t+1}, o_t, a_t, g) + \\ \beta_{o_t}(s_{t+1}) \max\limits_{a'} Q_U(s_{t+1}, o_t, a', g) \Bigr) + \Bigr] + \\ (1 - \alpha)Q_U(s_t, o_t, a_t, g)
\end{multline*}

We can assume $\forall s \in \mathcal{S}: \beta_{o_t}(s) = 1$, that is, all options terminate everywhere, to make the equations into a variation of Q-Learning:

\begin{multline*}
Q_\Omega(s_t, o_t, g) \leftarrow \alpha \Bigl[ r_t + \gamma \max\limits_{o'} Q_\Omega(s_{t+1}, o', g)  \Bigr] \\ + (1 - \alpha)Q_\Omega(s_t, o_t, g)
\end{multline*}

and 

\begin{multline*}
    Q_U(s_t, o_t, a_t, g) \leftarrow \alpha \Bigl[ r_t + \gamma  \max\limits_{a'} Q_U(s_{t+1}, o_t, a', g) + \Bigr] \\ + (1 - \alpha)Q_U(s_t, o_t, a_t, g)
\end{multline*}

Like \citet{schaul15uvfa}, we find that learning H-UVFAs via bootstrapping is highly unstable and prone to high variance. We were able to learn H-UVFAs using a low update rate. We were unable to learn UVFAs using bootstrapping, except for the option-value function which is equivalent to learning UVFAs with actions instead of options.

\section{Comparison with UVFAs} \label{uvfa-comparison}

Learning UVFAs, instead of H-UVFAs, would result in two two-dimensional matrices that are factored as

$$Q_\Omega(s, o, g; \theta) = h(\Phi(s, o), \Psi(g))$$
$$Q_U(s, o, a, g; \eta) = h(\phi(s, o, a), \psi(g))$$

Learning $Q_\Omega(s, o, g; \theta)$ is akin to learning UVFAs with options instead of actions; however, learning $Q_U(s, o, a, g; \eta)$ involves compressing a lot of information about the state, option, and action into a single embedding. This compression of a large amount of information into the single embedding leads to the poor performance and high variance of UVFAs as the policies/value functions of different options are not interchangeable or comparable by a similarity measure.

This is visualized in Appendix \ref{option_behaviors} where different options specialize to different rooms and generate behaviors pertaining to their "sub-goal" to achieve the larger goal. Since we get options that specialize to each room, it is not possible to use the value functions of an option specialised in one room to learn the behavior of an option specialized in another room. This is further bolstered in the reinforcement learning approach, since one would not acquire enough samples of all the options-action pair in a room where that option is barely picked.

As such, we draw the following comparisons between H-UVFAs and UVFAs in \textit{hierarchical settings}:
\begin{itemize}
    \item H-UVFAs achieve better performance and generalize better since they split all the embeddings into their respective dimensions instead of compressing a large amount of information into a single embedding. This is further applicable in the reinforcement learning methods of training H-UVFAs
    \item H-UVFAs require less data and are less computationally intense. In the reinforcement learning method one would require a large amount of data to explore more option-action pairs.
    \item H-UVFAs are more stable. This is seen in the reinforcement learning experiments where they have significantly lower variance in the Horde of targets approach and can be learned with the bootstrapping method for the lower-level value functions.
\end{itemize}



\section{Discussion and Future Directions}

We have introduced Hierarchical Universal Value Function Approximators (H-UVFAs), a universal approximator for goal-directed hierarchical tasks. We discuss supervised learning and reinforcement learning based approaches to learn H-UVFAs and provide experiments for the same. We show generalization of learned H-UVFAs to unseen tasks, and show that H-UVFAs have better performance and generalization to the baseline of UVFAs.

As a future direction, we propose a work that extends the universal approximator to also include the termination function used in hierarchical reinforcement learning. 


\section*{Acknowledgements}

We would like to thank Professor Eliot Moss and Ignacio Gavier for their insightful comments and discussions.

\section*{Impact Statement}

This paper presents work whose goal is to advance the field of 
Machine Learning. There are many potential societal consequences 
of our work, none which we feel must be specifically highlighted here.

{\raggedright
\bibliography{H-UVFA}

\begin{thebibliography}{11}
\providecommand{\natexlab}[1]{#1}
\providecommand{\url}[1]{\texttt{#1}}
\expandafter\ifx\csname urlstyle\endcsname\relax
  \providecommand{\doi}[1]{doi: #1}\else
  \providecommand{\doi}{doi: \begingroup \urlstyle{rm}\Url}\fi

\bibitem[{Bellemare} et~al.(2013){Bellemare}, {Naddaf}, {Veness}, and {Bowling}]{bellemare13arcade}
{Bellemare}, M.~G., {Naddaf}, Y., {Veness}, J., and {Bowling}, M.
\newblock The {Arcade} learning environment: {An} evaluation platform for general agents.
\newblock \emph{Journal of Artificial Intelligence Research}, 47:\penalty0 253--279, jun 2013.

\bibitem[Keshavan et~al.(2009)Keshavan, Oh, and Montanari]{keshavan09optspace}
Keshavan, R.~H., Oh, S., and Montanari, A.
\newblock Matrix completion from a few entries.
\newblock \emph{CoRR}, 2009.
\newblock \doi{abs/0901.3150}.

\bibitem[Kolda \& Bader(2009)Kolda and Bader]{kolda09parafac}
Kolda, T.~G. and Bader, B.~W.
\newblock Tensor decompositions and applications.
\newblock \emph{SIAM Review}, 51\penalty0 (3):\penalty0 455--500, 2009.
\newblock \doi{10.1137/07070111X}.
\newblock URL \url{https://doi.org/10.1137/07070111X}.

\bibitem[Precup(2000)]{precup00options}
Precup, D.
\newblock \emph{Temporal Abstraction in Reinforcement Learning}.
\newblock PhD thesis, University of Massachusetts Amherst, 2000.

\bibitem[Schaul et~al.(2015)Schaul, Horgan, Gregor, and Silver]{schaul15uvfa}
Schaul, T., Horgan, D., Gregor, K., and Silver, D.
\newblock Universal value function approximators.
\newblock In Bach, F. and Blei, D. (eds.), \emph{Proceedings of the 32nd International Conference on Machine Learning}, volume~37 of \emph{Proceedings of Machine Learning Research}, pp.\  1312--1320, Lille, France, 07--09 Jul 2015. PMLR.
\newblock URL \url{https://proceedings.mlr.press/v37/schaul15.html}.

\bibitem[Silva et~al.(2012)Silva, Konidaris, and Barto]{dasilva2012skills}
Silva, B.~D., Konidaris, G., and Barto, A.
\newblock Learning parameterized skills.
\newblock In \emph{Proceedings of the 29th International Conference on Machine Learning 2012}, 2012.
\newblock URL \url{https://arxiv.org/abs/1206.6398}.

\bibitem[Sutton \& Barto(1998)Sutton and Barto]{barto98rlbook}
Sutton, R.~S. and Barto, A.~G.
\newblock \emph{Reinforcement Learning: {An} Introduction}.
\newblock A Bradford Book, Cambridge, MA, USA, 1998.
\newblock ISBN 0262039249.

\bibitem[Sutton et~al.(1999)Sutton, Precup, and Singh]{sutton99smdp}
Sutton, R.~S., Precup, D., and Singh, S.
\newblock Between {MDPs} and semi-{MDPs}: {A} framework for temporal abstraction in reinforcement learning.
\newblock \emph{Artificial Intelligence}, 112\penalty0 (1):\penalty0 181--211, 1999.
\newblock ISSN 0004-3702.
\newblock \doi{https://doi.org/10.1016/S0004-3702(99)00052-1}.
\newblock URL \url{https://www.sciencedirect.com/science/article/pii/S0004370299000521}.

\bibitem[Sutton et~al.(2011)Sutton, Modayil, Delp, Degris, Pilarski, White, and Precup]{sutton11gvf}
Sutton, R.~S., Modayil, J., Delp, M., Degris, T., Pilarski, P.~M., White, A., and Precup, D.
\newblock Horde: {A} scalable real-time architecture for learning knowledge from unsupervised sensorimotor interaction.
\newblock In \emph{The 10th International Conference on Autonomous Agents and Multiagent Systems - Volume 2}, AAMAS '11, pp.\  761–768, Richland, SC, 2011. International Foundation for Autonomous Agents and Multiagent Systems.
\newblock ISBN 0982657161.

\bibitem[Tomasi \& Bro(2005)Tomasi and Bro]{tomasi05parafac}
Tomasi, G. and Bro, R.
\newblock {PARAFAC} and missing values.
\newblock \emph{Chemometrics and Intelligent Laboratory Systems}, 75:\penalty0 163--180, 02 2005.
\newblock \doi{10.1016/j.chemolab.2004.07.003}.

\bibitem[Tucker(1966)]{tucker66decomposition}
Tucker, L.~R.
\newblock Some mathematical notes on three-mode factor analysis.
\newblock \emph{Psychometrika}, 31:\penalty0 279–311, 1966.
\newblock \doi{https://doi.org/10.1007/BF02289464}.

\end{thebibliography}
\bibliographystyle{icml2024}
}

\newpage
\appendix
\onecolumn
\section{Experiment Details}  \label{experiment_details_appendix}

We use the Four Rooms domain \cite{sutton99smdp} for our experiments. The agent can be initialized in any of the four rooms. During training, the goal can be initialized in only three of the four rooms. When testing for generalization, the goal is picked only in the fourth room. The agent received a reward of 1 for reaching the goal and 0 otherwise. The episodes terminate after 1000 timesteps.

When evaluating H-UVFAs and UVFAs, we assume the agent can terminate in a given state with a probability of 0.5. When training H-UVFAs and UVFAs, we assume the agent terminates in one step when applicable. All networks we use consist of hidden layers of size 128 and Rectified Linear Units activation functions. We use a learning rate of 0.05 for the supervised learning approach and reinforcement learning Horde approach. We also experimented with other learning rates. The states and goals are trained with a batch size of 16, while the options and actions are trained with a batch size of 2.

\textbf{Atari: } The Ms. Pacman experiments are performed on a modified environment where each pellet represents a goal. The training set consists of 30 pellets and the test set for generalization consists of the remaining 120 pellets. Ms. Pacman is initialized at a random position at the start of each episode. A PyTorch implementation of the Option-Critic architecture is used to train each agent in the Horde. The inputs are grayscale images processed with a convolutional network.

\section{Factorized Matrices and Re-creations} \label{matrix_recreations}

\begin{figure}[h]
\vskip 0.2in
\begin{center}
\centerline{\includegraphics[width=0.5\columnwidth]{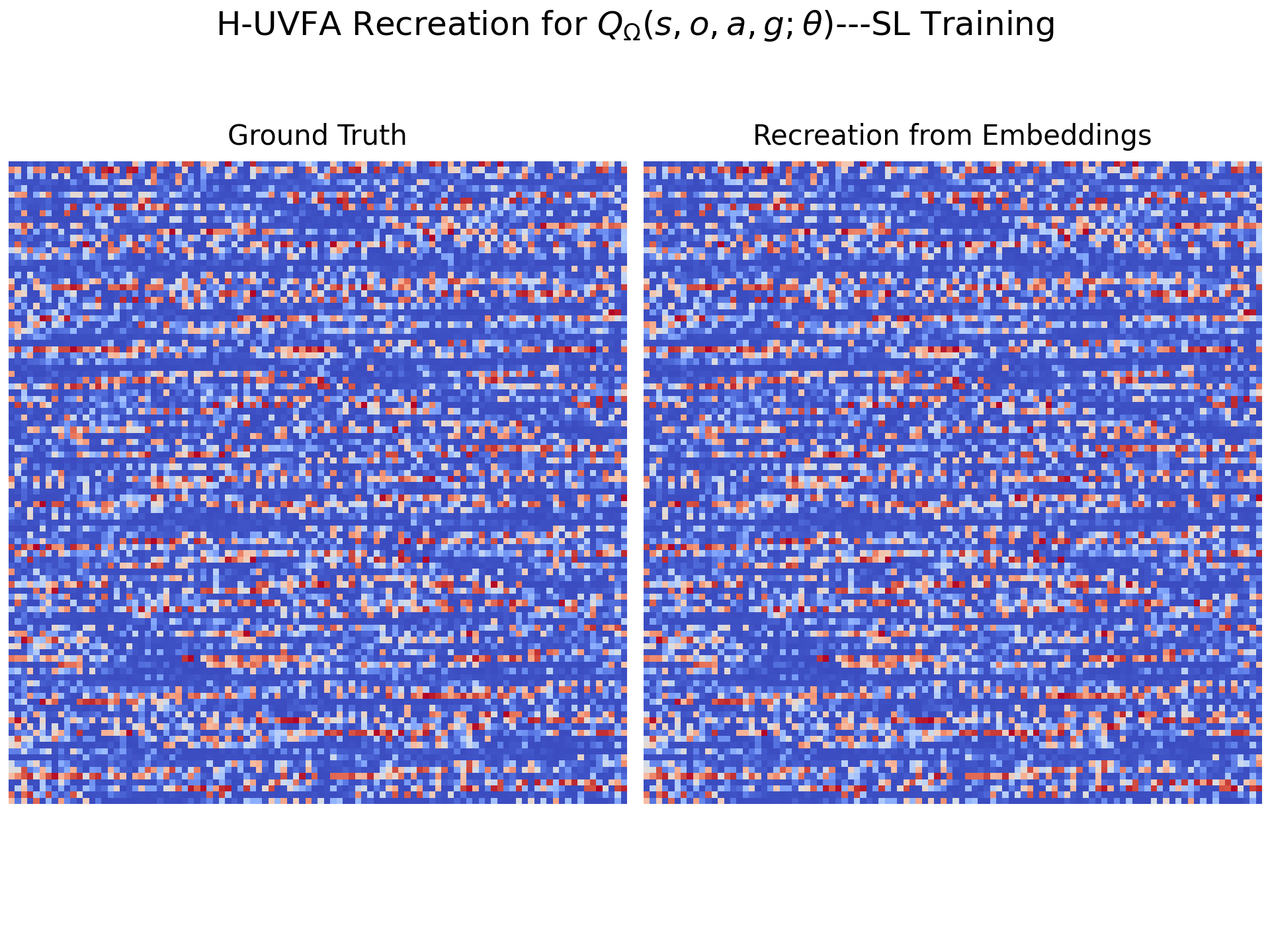}}
\centerline{\includegraphics[width=0.5\columnwidth]{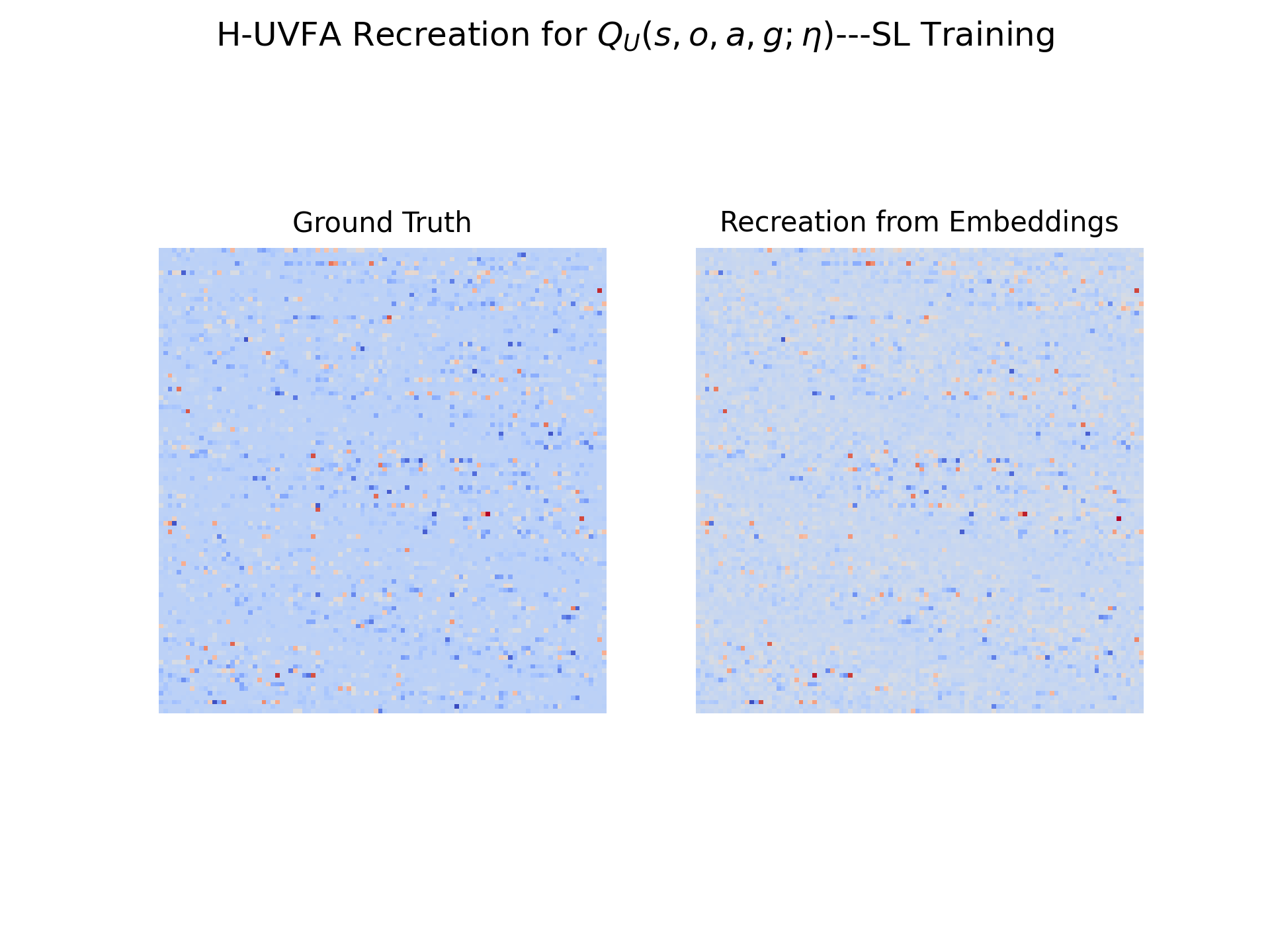}}
\caption{A comparison of the ground truth and H-UVFA for $Q_\Omega(s, o, g; \theta)$ and $Q_U(s, o, a, g; \eta)$ when H-UVFAs are created using supervised learning.}
\label{fig:appendix-sl-recreation}
\end{center}
\vskip -0.2in
\end{figure}

$$$$

\begin{figure}[h]
\vskip 0.2in
\begin{center}
\centerline{\includegraphics[width=0.5\columnwidth]{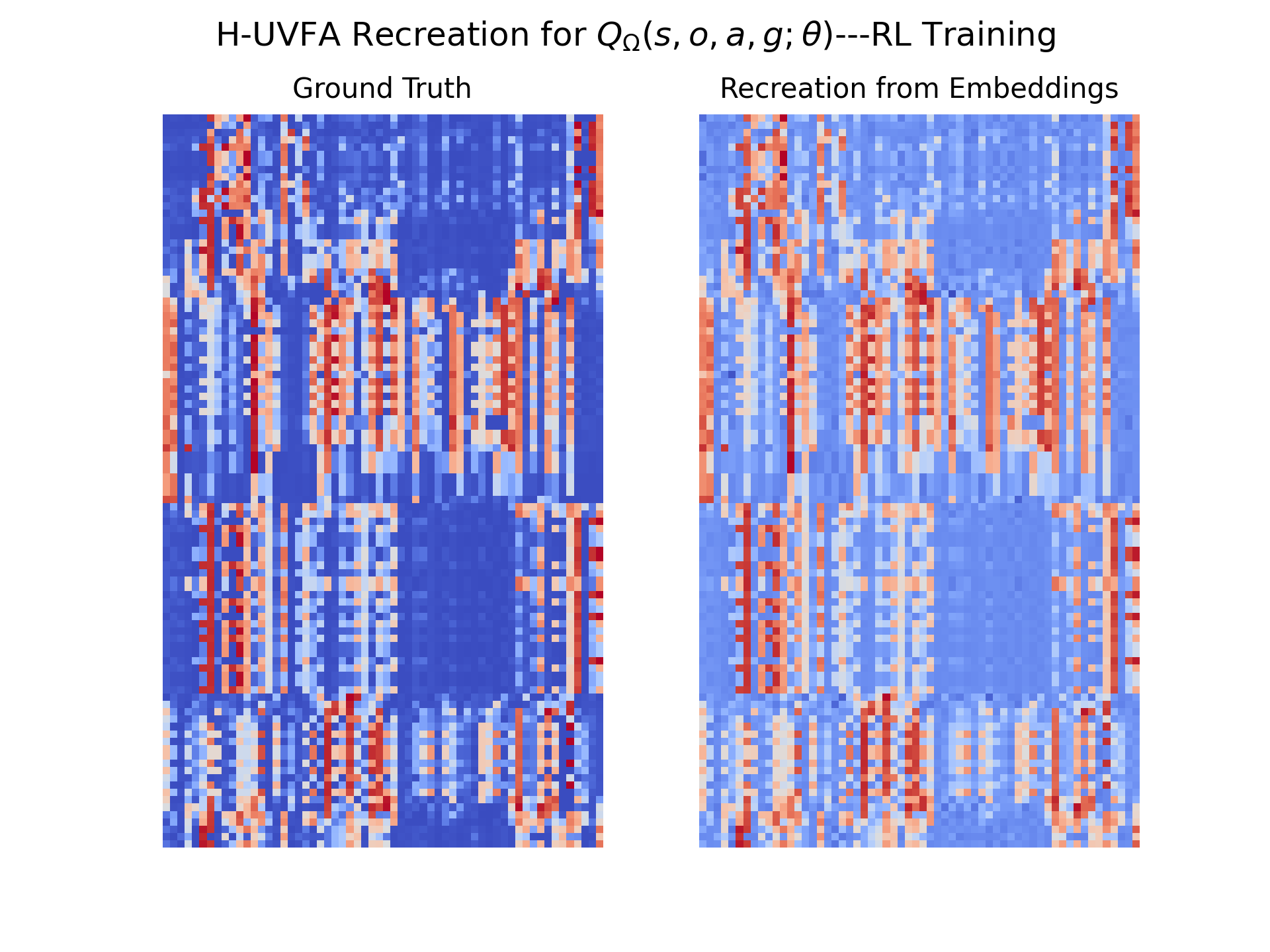}}
\centerline{\includegraphics[width=0.5\columnwidth]{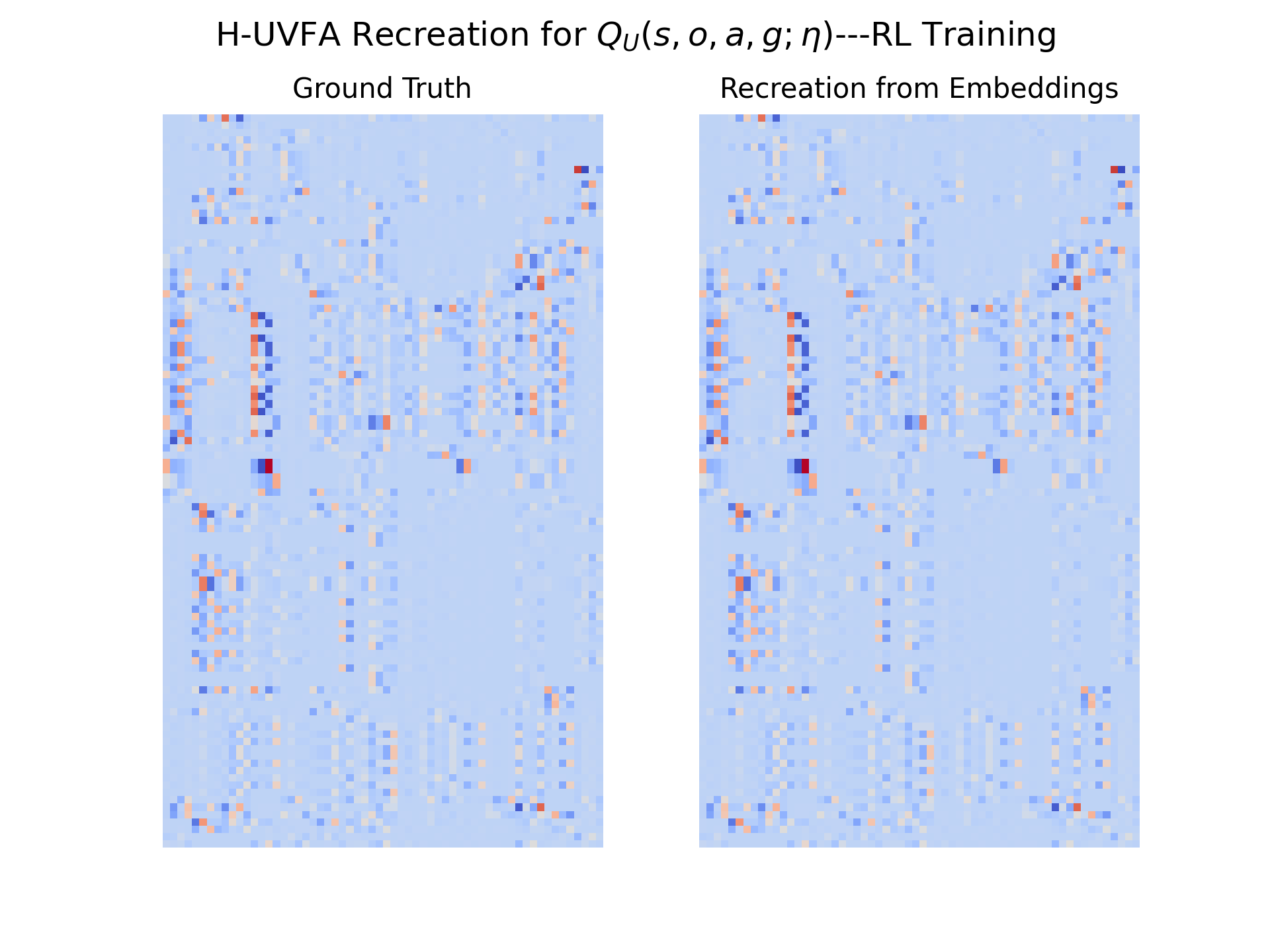}}
\caption{A comparison of the ground truth and H-UVFA for $Q_\Omega(s, o, g; \theta)$ and $Q_U(s, o, a, g; \eta)$ when H-UVFAs are created using the reinforcement learning.}
\label{fig:appendix-rl-recreation}
\end{center}
\vskip -0.2in
\end{figure}

$$$$

\newpage
\begin{figure}[!h]
\vskip 0.2in
\begin{center}
\centerline{\includegraphics[width=0.9\columnwidth]{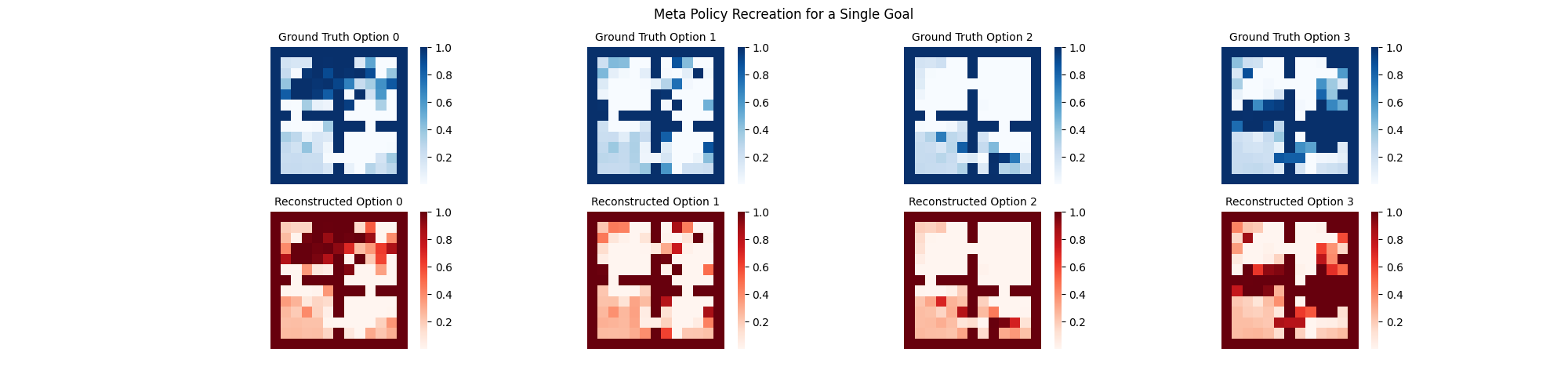}}
\centerline{\includegraphics[width=0.9\columnwidth]{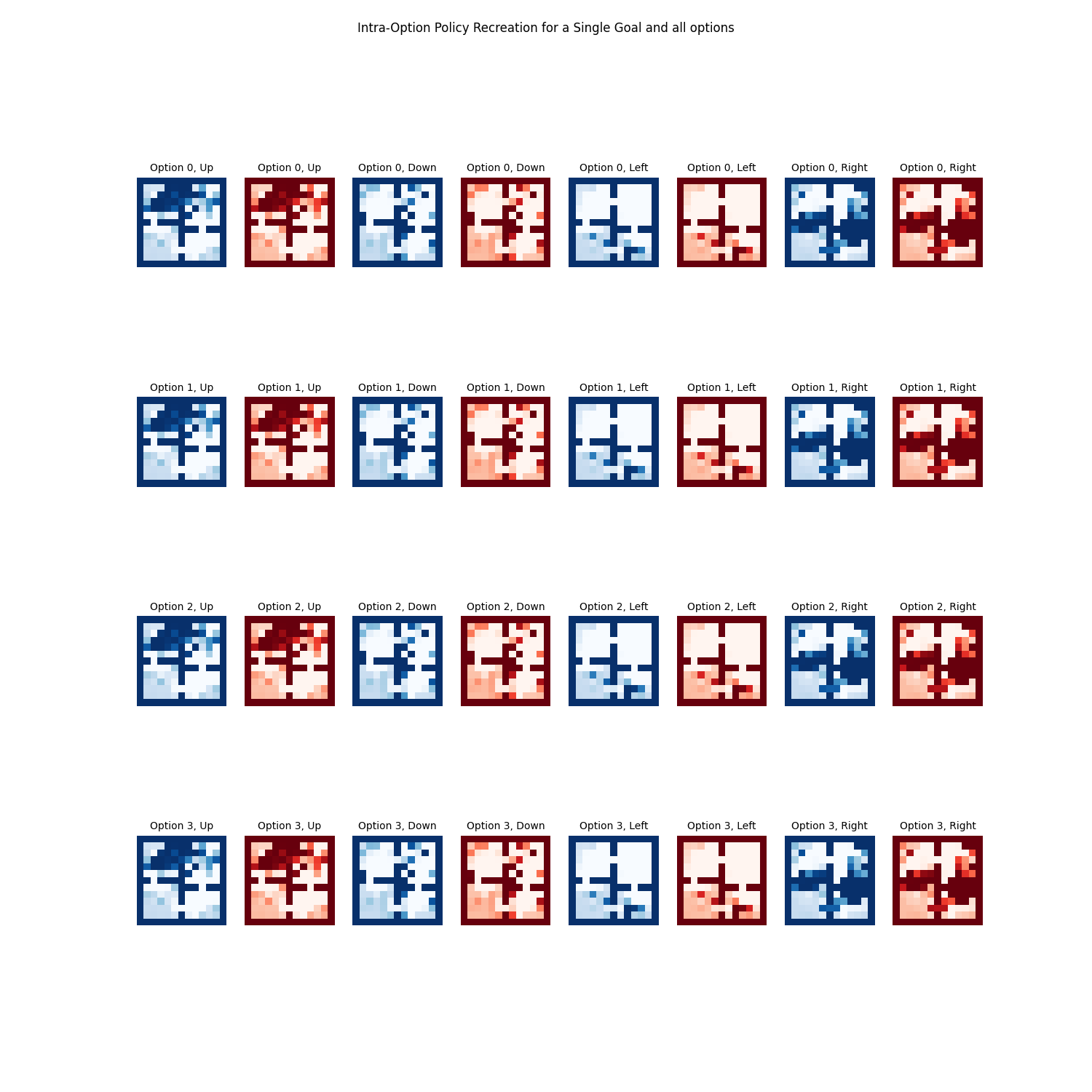}}
\caption{The re-creation of the meta-policy and option-policies for a single goal with the embeddings trained on 25 goals. The ground truth is in blue with the re-creations in red.}
\label{fig:appendix-recreation}
\end{center}
\vskip -0.2in
\end{figure}

$$$$

\newpage
\section{Explaining UVFA Performance: Options as Specialized Behaviors} \label{option_behaviors}

\begin{figure}[!h]
\vskip 0.2in
\begin{center}
\centerline{\includegraphics[width=0.9\columnwidth]{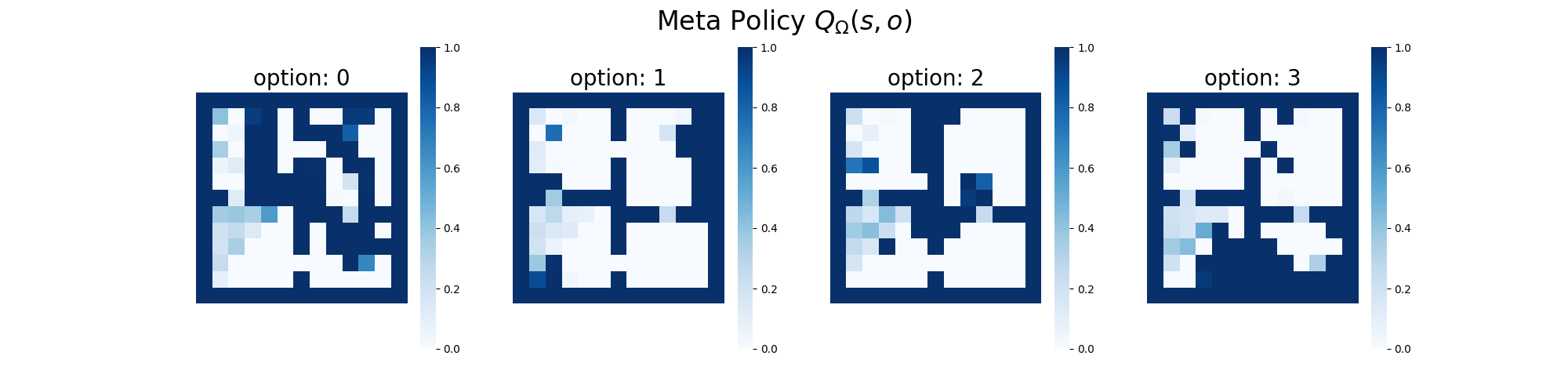}}
\centerline{\includegraphics[width=0.69\columnwidth]{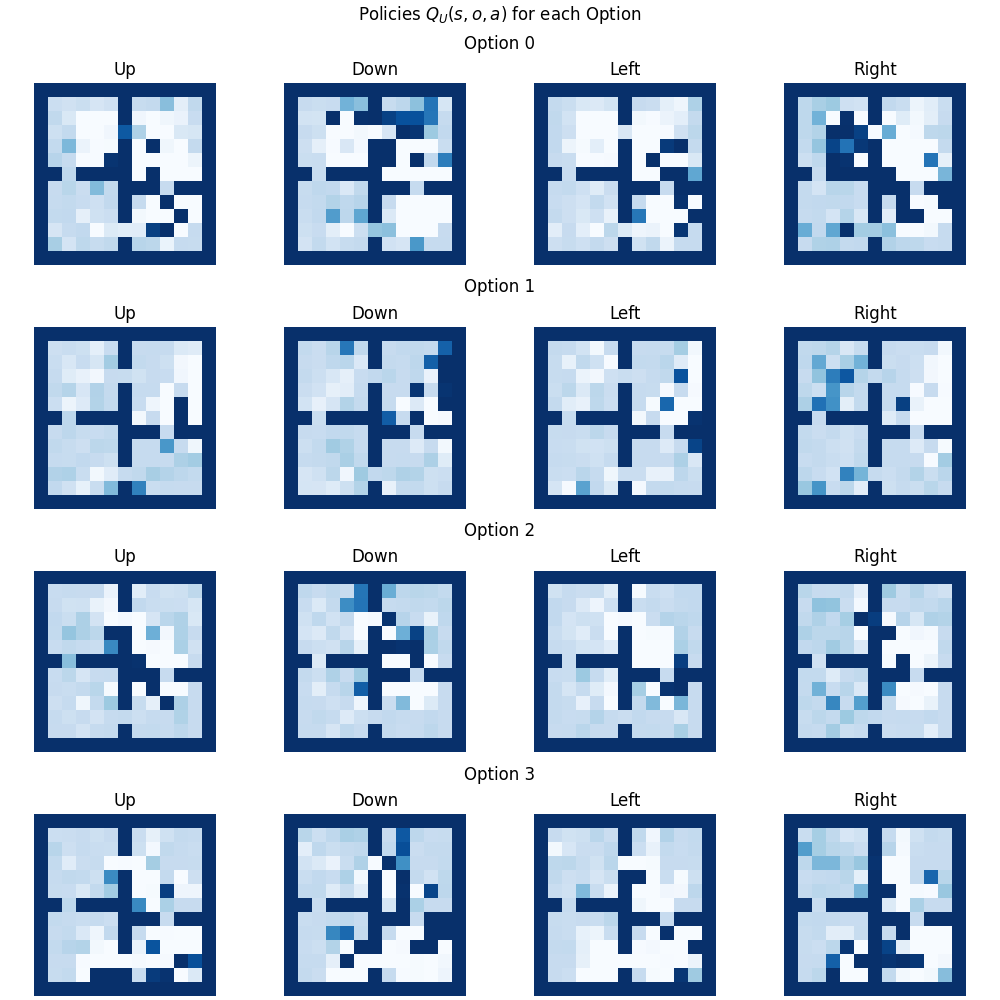}}
\caption{The Policy-over-options and intra-option policies for a single goal and four options when the goal is located right below the right corridor.}
\label{fig:appendix-single-goal}
\end{center}
\vskip -0.2in
\end{figure}

Figure \ref{fig:appendix-single-goal} plots the meta-policy and intra-option policies when the goal is located right below the right corridor.

We can see that the options specialize to a subset of the rooms and do not learn a policy in other rooms. The policy of the option in a room it doesn't initialize in leads to sub-optimal or even incorrect behavior. For instance, option 0 doesn't initialize in the bottom-left room, option 1 doesn't initialize in the bottom-right room and option 3 doesn't initialize in the top-right room. The policies of the options in these rooms is non-existent and even moves the agent away from the goal.

Hence, these options learn specialized skills and behaviors pertaining to a "subgoal" construct. In the reinforcement learning of UVFAs, the history might only include transitions pertaining to a given option in a given area of the map, however, that option might be specialized to another room for another goal and hence the relevant information is not learned by the UVFA. Using H-UVFAs gives us the granular control needed to select the correct options in the necessary rooms.

\end{document}